\documentclass[journal]{IEEEtran}
\usepackage{amsmath,amsfonts}
\usepackage{algorithm}
\usepackage{algpseudocode}
\usepackage{array}
\usepackage{textcomp}
\usepackage{stfloats}
\usepackage{url}
\usepackage{xurl}
\usepackage{verbatim}
\usepackage{graphicx}
\usepackage{cite}
\usepackage{xcolor}
\usepackage{tabularx}
\usepackage{booktabs}
\usepackage{colortbl} 
\usepackage{placeins}
\usepackage{multirow}
\begin{document}

\title{CoRe-Net: Co-Operational Regressor Network with Progressive Transfer Learning for Blind Radar Signal Restoration}

\author{Muhammad Uzair Zahid, Serkan Kiranyaz,~\IEEEmembership{Senior Member,~IEEE}, Alper Yildirim,~\IEEEmembership{Senior Member,~IEEE},\\and Moncef Gabbouj,~\IEEEmembership{Fellow,~IEEE}
\thanks{Muhammad Uzair Zahid is with the Department of Computing Sciences, Tampere University, 33100 Tampere, Finland (e-mail: muhammaduzair.zahid@tuni.fi).}
\thanks{Serkan Kiranyaz is with the Department of Electrical Engineering,
College of Engineering, Qatar University, Qatar.}
\thanks{Alper Yildirim is with the Department of Electrical and Electronics Engineering, School Of Engineering and Natural Sciences, Ankara Medipol University, Turkey.}
\thanks{Moncef Gabbouj is with the Department of Computing Sciences, Tampere University, Finland.}}

\markboth{Journal of \LaTeX\ Class Files,~Vol.~14, No.~8, August~2021}%
{Shell \MakeLowercase{\textit{et al.}}: A Sample Article Using IEEEtran.cls for IEEE Journals}


\maketitle

\begin{abstract}
Real-world radar signals are frequently corrupted by various artifacts, including sensor noise, echoes, interference, and intentional jamming, differing in type, severity, and duration. Traditional radar signal restoration methods often address a single restoration problem, such as a specific noise type, and typically assume a fixed signal-to-noise ratio (SNR) level. This severely limits their usability to practical real-world applications. Several Deep Learning models have been proposed for restoration and recently Operational Generative Adversarial Networks (OpGANs) have achieved state-of-the-art performance levels. However, adversarial-based methods may still face instability and challenges with limited restoration quality refinement. This pilot study introduces a novel model, called Co-Operational Regressor Network (CoRe-Net) for blind radar signal restoration, designed to address such limitations and drawbacks. CoRe-Net replaces adversarial training with a novel cooperative learning strategy, leveraging the complementary roles of its Apprentice Regressor (AR) and Master Regressor (MR). The AR restores radar signals corrupted by various artifacts, while the MR evaluates the quality of the restoration and provides immediate and task-specific feedback, ensuring stable and efficient learning. The AR, therefore, has the advantage of both self-learning and assistive learning by the MR. The proposed model has been extensively evaluated over the benchmark Blind Radar Signal Restoration (BRSR) dataset, which simulates diverse real-world artifact scenarios. Under the fair experimental setup, this study shows that the CoRe-Net surpasses the Op-GANs over a 1 dB mean SNR improvement. To further boost the performance gain, this study proposes multi-pass restoration by cascaded CoRe-Nets trained with a novel paradigm called  Progressive Transfer Learning (PTL), which enables iterative refinement, thus achieving an additional 2 dB mean SNR enhancement. Multi-pass CoRe-Net training by PTL consistently yields incremental performance improvements through successive restoration passes whilst highlighting CoRe-Net’s ability to handle such a complex and varying blend of artifacts. As a result, CoRe-Net sets a new benchmark for blind radar signal restoration by achieving state-of-the-art performance with a significant gap over a wide range of SNR levels of the input signals corrupted by various artifact types. 
\end{abstract}


\section{Introduction}
\IEEEPARstart{R}{adar signal} restoration is a crucial process that enhances the accuracy and reliability of radar systems by mitigating the effects of noise and interference arising from atmospheric conditions, hardware imperfections, and external sources. Radar systems can detect and identify targets with greater precision by effectively restoring the signal received. This is particularly important for various applications where the reliability of radar systems is of utmost importance, such as weather forecasting, target tracking, and automotive safety. In the domain of electronic warfare (EW), accurately detecting and classifying radar signals received from a possible target at the longest possible distance is especially important. However, EW receivers often operate across wider frequency bands and are exposed to higher noise levels, resulting in lower SNR, which hinders accurate detection. For instance, reliable measurements of instantaneous signal parameters, such as direction of arrival (DOA), typically require an SNR exceeding \( 18 \, \text{dB} \) \cite{de2018introduction}. Enhancing the SNR in such cases is vital for improving the detection and classification capabilities of EW sensors. Radar systems in general also face challenges in maintaining reliable detection under noisy conditions. Detection theory suggests that achieving a probability of detection (\( P_d = 90\% \)) and a low probability of false alarm (\( P_{fa} = 10^{-6} \)) requires an SNR of at least \( 13 \, \text{dB} \) for nonfluctuating signals reflected from targets \cite{tsui1986microwave}. To increase the SNR level, traditional radar systems often integrate multiple pulses, consuming additional resources and time to improve detection accuracy. Low probability of intercept (LPI) radars present a unique challenge due to their low output power and their ability to distribute energy over a wide frequency spectrum, making them more difficult to detect \cite{pace2009detecting}. To overcome these challenges, enhancing the SNR, also known as signal processing gain, plays an important role in the detection of weak signals received from long distances. Radar signal restoration continues to be an important component of radar signal processing, as well as a focus of research in remote sensing and radar system design.

\begin{figure*}[t!]
    \centering
    \includegraphics[width=1\textwidth,keepaspectratio]{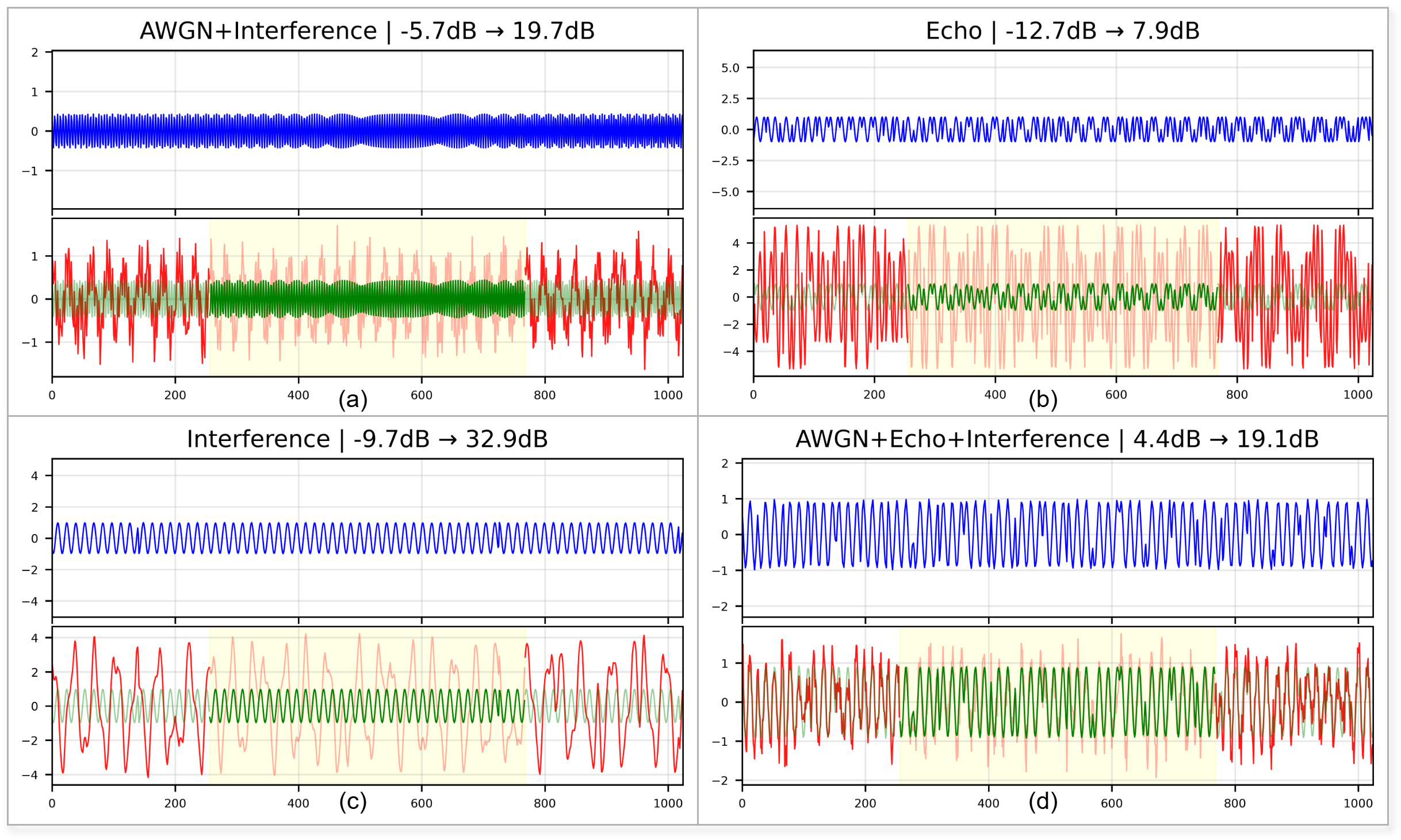}
    \caption{Sample signals from the BRSR dataset demonstrating the impact of various artifacts and the effectiveness of the proposed Core-Net approach. In each example, the clean signal (blue) is shown in the top plot, whereas the corrupted (red) and restored (green) signals are overlaid in the bottom plot. The middle area in the bottom plot is highlighted to emphasize the restored signal. Artifact combinations include AWGN + Interference (a), Echo (b), Interference (c), and a blend of all artifacts (d), with annotated SNR improvements showcasing the restoration performance.}
    \label{fig:intro_artifacts}
\end{figure*}

As the number of radar radiation sources continues to grow, the electromagnetic environment becomes increasingly complex, leading to significant challenges in radar signal processing. This complexity often introduces severe distortions in the received radar signals, resulting in a low SNR that complicates accurate signal recognition and classification. Traditional radar signal restoration methods generally focus on addressing specific distortion types under fixed SNR conditions, tackling isolated tasks such as denoising or echo reduction. While these approaches demonstrate effectiveness in controlled or idealized scenarios, they often fall short when faced with the multifaceted and dynamic challenges of real-world environments, where distortions are unpredictable, blended, and diverse in nature.

In a typical real-world scenario, various artifacts from different corruption schemes, such as (additive) sensorial noise, echo, and interference, are encountered with various types, severities, and durations, as illustrated in Fig.~\ref{fig:intro_artifacts}. The complex and dynamic nature of such distortions necessitates a comprehensive and blind restoration approach capable of handling a wide spectrum of signal corruption scenarios encountered in real-world applications. In essence, such an approach should go beyond the limitations of conventional methods and provide robust solutions that can adapt to the diverse and dynamic nature of radar signals in practice.

Time-frequency analysis is an important technique for extracting intra-pulse features of radar emitters from their signals by transforming them into time-frequency images (TFIs). Widely used methods, such as the Short-Time Fourier Transform (STFT) \cite{zhou2019ensemble}, Wigner-Ville Distribution (WVD) \cite{kishore2017automatic}, and Choi-Williams Distribution (CWD) \cite{seddighi2020radar}, provide detailed representations of signals across both time and frequency domains. These representations serve as the basis for advanced feature extraction and waveform identification. Building on these transformations, researchers have increasingly incorporated deep learning (DL) techniques for automated analysis of TFIs, leading to improved feature extraction and waveform identification \cite{jan2020artificial,zhou2018automatic,zhang2017convolutional}. A notable example is the multilayer denoising approach proposed in \cite{jiang2024multilayer}, which applies variational mode decomposition (VMD), local mean decomposition (LMD), and wavelet thresholding (WT) in sequence to decompose noisy radar signals. The processed signals are subsequently converted into TFIs via the CWD. A neural network employing dilated convolution is trained over them, yielding a 75.3\% recognition rate under low SNR conditions. Similarly, \cite{lee2019mutual} presents a wavelet-based denoising strategy for mitigating mutual interference in automotive FMCW radar systems, demonstrating marked improvements in target detection accuracy.

Recent developments in deep learning have opened new avenues for radar signal denoising. For example, Fuchs et al. \cite{fuchs2020automotive} proposed an interference mitigation scheme using a convolutional autoencoder for automotive radar applications, underscoring the effectiveness of deep learning in enhancing the signal-to-noise-plus-interference ratio. Meanwhile, other studies \cite{wang2017automatic,kong2018automatic,zhang2019automatic} rely on time-frequency analysis to convert raw radar signals into 2D images, which are then processed by 2D deep convolutional neural networks (CNNs), a workflow that often incurs high computational costs.
In an effort to streamline both denoising and classification, Du et al. \cite{du2022dncnet} introduced DNCNet, a deep model with two distinct subnets: one for denoising using a U-Net architecture enhanced by a noise-level estimation module and another for classification. Their two-phase training procedure first focuses on optimizing the denoising subnetwork, followed by a stage aimed at refining the mapping between the denoised signals and their perceptual representations. Although the study does not directly evaluate the quality of the restored signals, its emphasis on classification performance under varying SNR conditions demonstrates that substantial gains are possible by robust denoising.

GANs \cite{goodfellow2014generative} have revolutionized signal processing by providing powerful frameworks to tackle complex restoration tasks \cite{mvuh2024multichannel}. Through their adversarial training process, GANs have achieved remarkable success in various domains ranging from image enhancement to speech denoising and are increasingly being explored for radar signal restoration. An important and recent evolution of GAN architectures, known as Operational GANs (OpGANs), incorporates 1D Self-Organized Operational Neural Network (Self-ONN)\cite{kiranyaz2021self,zahid2021robust,gabbouj2022robust,zahid2022global} layers into the generator, enabling robust mappings from degraded to clean signals while accommodating a broader range of real-world artifacts \cite{ince2022blind,kiranyaz2022blind}.  Numerous prior studies have shown that 1D and 2D Self-ONNs can outperform conventional CNNs in various regression and classification tasks due to their efficient computational design and generative neuron model. In \cite{zahid2024brsr}, ``Blind Radar Signal Restoration'' framework using OpGANs (BRSR-OpGAN) has recently been proposed and achieved state-of-the-art restoration performance with a significant gap. Notably, the BRSR-OpGAN framework employs compact 1D Self-ONNs in both generator and discriminator networks to perform blind radar signal restoration without assuming any prior knowledge of artifact types or severity. Although the generator and discriminator have only 11 and 7 layers, respectively, BRSR-OpGAN achieves such a superior restoration by optimizing the process in both time and frequency domains via a dual-domain loss function. Evaluated on the BRSR dataset, a benchmark simulating realistic radar conditions, BRSR-OpGAN has demonstrated unprecedented blind restoration performance across multiple distortion scenarios, affirming the efficacy of integrating Self-ONN layers within GAN-based approaches for radar signal processing.

Despite its success, BRSR-OpGAN, similar to other GAN-based methods, encountered challenges related to adversarial training instability and limited capacity for iterative refinement. Inspired by the R2C-GAN framework \cite{ahishali2024r2c}, where integrating a classification task into the generative adversarial network improved restoration and diagnostic performance, this study explores how task-specific feedback could further enhance restoration performance. While R2C-GANs demonstrated the effectiveness of coupling restoration with the classification of the medical images, the proposed CoRe-Net model generalizes this concept by further introducing a collaborative training paradigm. Instead of adversarial training, the CoRe-Net leverages task-specific feedback through a Master Regressor (MR) that predicts a quantitative metric (e.g., SNR) to assist the training of the Apprentice Regressor (AR) similar to how a teacher improves a student's learning process by a stream of continuous assessment feedbacks. 

Building on this foundation, this study shows that CoRe-Nets can address the limitations of adversarial methods by adopting a cooperative learning strategy, which aligns task-specific feedback with restoration objectives, ensuring stability and efficiency in the ongoing training process. Furthermore, using Progressive Transfer Learning (PTL) to perform multiple training passes enables significant performance improvements by leveraging the restored (output) signal from one pass as the input for the next pass. After the first pass, each CoRe-Net inherits the network parameters of the previous pass to initiate its training, and thus, such an internal PTL mechanism benefits from the accumulated learning in the prior passes and thus yields significant performance gains in each pass.  Through comprehensive testing and analysis, this research showcases the potential of CoRe-Nets with PTL for blind signal restoration and can also be used for other challenging signal-to-signal regression tasks. Overall, the novel and major contributions of this study can be summarized as follows:

\begin{enumerate}
  \item This study introduces a novel approach for blind radar signal restoration, where no prior assumptions are made about the types or severity of artifacts corrupting the radar signal.
  \item It proposes a pioneer Co-operational Regressor Network model, the CoRe-Net, leveraging AR and MR to collaboratively improve the restoration performance.
  \item This study further incorporates the Progressive Transfer Learning (PTL) paradigm, enabling iterative refinement of signal quality across multiple training passes, achieving significant SNR gains.
  \item The effectiveness of the proposed method is demonstrated on the benchmark \textit{BRSR Dataset}, which includes paired clean and corrupted signals simulating real-world radar environments with artifact severities spanning a wide SNR range of \(-14 \, \text{dB}\) to \(10 \, \text{dB}\).
  \item A detailed set of comparative evaluations reveals that the CoRe-Net can surpass the previous state-of-the-art model, BRSR-OpGAN, by a mean SNR improvement of more than \( 1 \, \text{dB} \). The gap widens beyond \( 3 \, \text{dB} \) by the multiple restoration passes with PTL. 
  \item The compact and computationally efficient AR model by 1D Self-ONNs in Core-Nets allows multi-pass radar signal restoration in real-time, even on resource-constrained hardware, making it suitable for practical applications.
\end{enumerate}

The rest of the paper is organized as follows:  
Section \ref{sec:methodology} introduces the proposed CoRe-Net framework, elaborating on its cooperative regression paradigm, the roles of the Apprentice and Master Regressors, the Progressive Transfer Learning (PTL) mechanism, and the architectural design of CoRe-Net.
Section \ref{sec:dataset} provides a brief overview of artifact modeling and the BRSR dataset, which simulates real-world radar signal conditions with diverse artifacts. It also details the data normalization, ensuring consistent preprocessing for robust model training and testing.
Section \ref{sec:results} presents the experimental setup and a detailed set of comparative evaluations, including quantitative results such as mean SNR and qualitative analyses of restored signals and comparisons against the competing methods. The section also highlights the impact of PTL on multi-pass restoration performance. Finally, the section concludes with a computational complexity analysis. 
Section \ref{sec:conclusion} summarizes the contributions of this study, discusses its implications for radar signal restoration, and identifies potential directions for future research.

\section{Methodology}
\label{sec:methodology}
This section presents the proposed CoRe-Net framework for blind radar signal restoration. We begin by formulating the problem by clearly defining the inputs, outputs, and target labels. We then introduce the 1D Co-Operational Regressor Network (CoRe-Net), comprised of an Apprentice Regressor (AR) and a Master Regressor (MR). The training dynamics, which describe how AR and MR can jointly be optimized in a cooperative manner, are subsequently discussed. Following this, we describe the multi-pass restoration approach by the ARs of the CoRe-Nets each of which is trained by Progressive Transfer Learning (PTL). Finally we provide detailed overview of Core-Net architectures, highlighting the design of AR and MR.

\subsection{Problem Formulation}
Real-world radar signals are often corrupted by numerous artifacts, including Additive White Gaussian Noise (AWGN), echo, and interference from other sources. These distortions can be mathematically expressed as:
\begin{equation}
r(t) = s(t) + n(t)
\label{eq:corrupted_signal}
\end{equation}
where \( r(t) \) is the distorted signal, \( s(t) \) is the clean radar signal, and \( n(t) \) represents an unknown combination of distortions. The characteristics of these distortions, along with their specific implementations in the dataset, are described in Section~\ref{sec:dataset}.

The objective of blind radar signal restoration is to estimate \( \hat{s} \) such that;
\(
\hat{s} \approx s,
\)
using only the corrupted input \( r \), without any prior knowledge about the artifact type, severity, or the overall SNR. This task is particularly challenging due to the diversity and unpredictability of the distortions and their combined effects.

\begin{figure*}[t!]
    \centering  \includegraphics[width=1\textwidth,keepaspectratio]{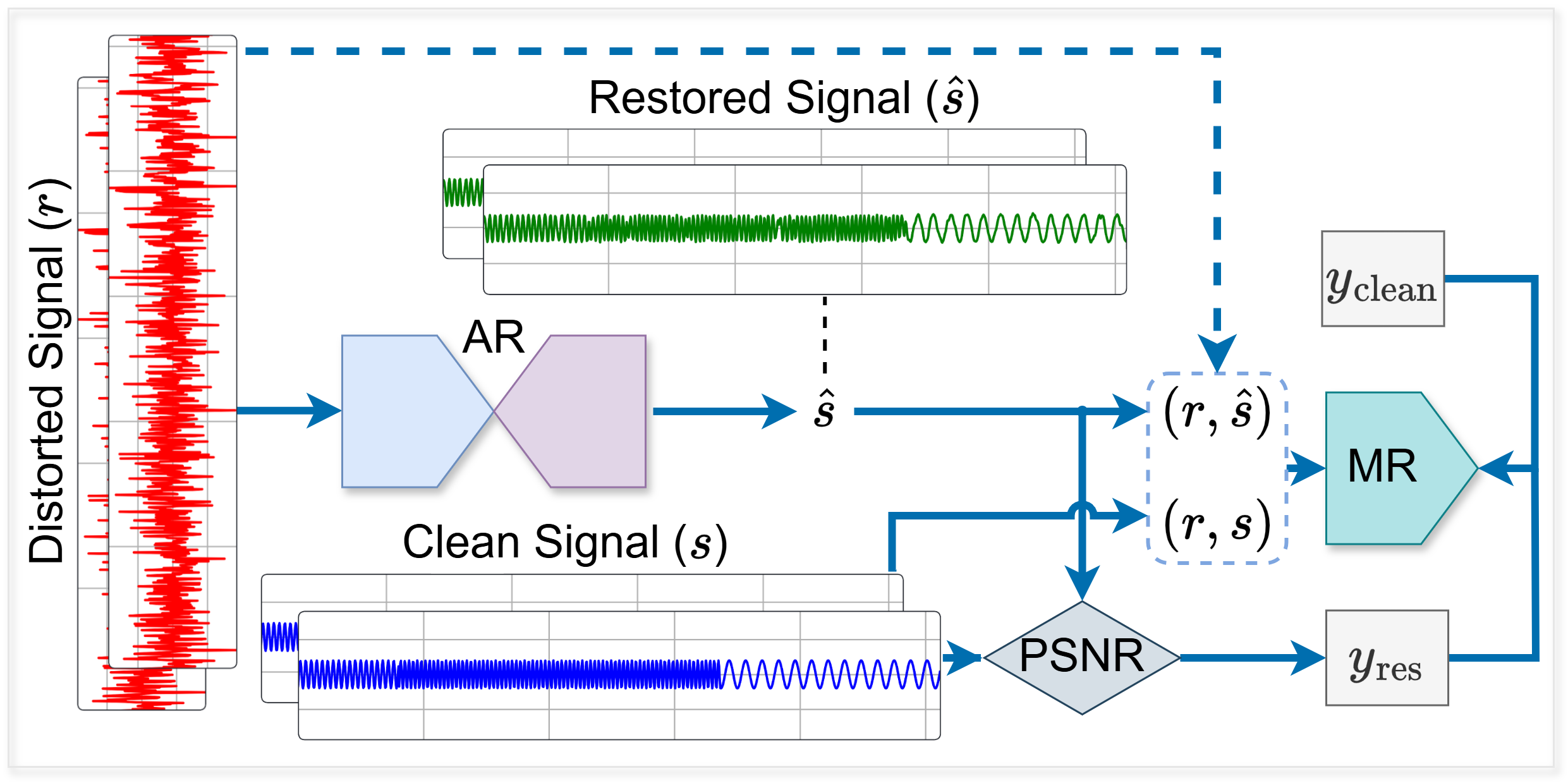}
    \caption{Overall methodology illustrating the CoRe-Net framework, where the AR restores the distorted radar signal (\(\hat{s}\)), and the MR predicts the PSNR for both restored and clean signals.}
    \label{fig:methodology}
\end{figure*}

\subsection{CoRe-Net}
\label{subsection:corenet}
The proposed CoRe-Net model is based on a cooperative learning paradigm, as illustrated in Fig.~\ref{fig:methodology}. Unlike traditional adversarial models such as GANs and their variants, CoRe-Net introduces a cooperative learning strategy as an alternative to adversarial training, leveraging the complementary roles of an Apprentice Regressor (AR) and a Master Regressor (MR). The AR is responsible for restoring radar signals that have been degraded by various artifacts. Concurrently, the MR assesses the restoration quality and delivers immediate, task-specific feedback to AR, thus promotes stable and efficient learning. This approach allows the AR to benefit from both autonomous and assistive (guided) learning facilitated by the MR. 

\paragraph{Apprentice Regressor (AR)}  
The AR is designed as a conditional signal regressor model \( A : r \mapsto \hat{s} \), where \(\hat{s}\) is the restored signal generated from the distorted input \( r \). Implemented as an encoder-decoder network, the AR minimizes the discrepancy between \( \hat{s} \) and the clean signal \( s \), leveraging a combination of fidelity, time-domain, and frequency-domain loss components. These losses, as detailed in Section~\ref{subsection:cooperative_learning_process}, ensure that the AR restores both temporal and spectral properties of the radar signal.

\paragraph{Master Regressor (MR)}  
The MR serves as a quality evaluator \( M \) of the AR's output, predicting a normalized PSNR score to assess the restored signals. For a clean input-output pair \((r, s)\), the MR outputs a value near \(1\), reflecting the ideal PSNR. For a restored pair \((r,\hat{s})\), the MR predicts the normalized PSNR \(y_{\text{res}} \in [0,1]\), computed as:

\begin{equation}
y_{\text{res}} = \frac{\mathrm{PSNR}(s, \hat{s})}{\mathrm{PSNR}_{\text{target}}} 
\label{eq:y_res}
\end{equation}

where, \(\mathrm{PSNR}_{\text{target}}\) is a predefined maximum achievable PSNR and \(\mathrm{PSNR}(\hat{s}, s)\) is expressed by:

\begin{equation}
\mathrm{PSNR}(\hat{s}, s) = 10 \log_{10} \left( \frac{\bigl(\max(s)\bigr)^2}{\mathrm{MSE}(s, \hat{s})} \right)
\label{eq:psnr_correct}
\end{equation}

where \(\max(s)\) represents the maximum possible value of the clean signal \(s\), and the Mean Squared Error (MSE) between \(s\) and \(\hat{s}\) is defined as:

\begin{equation}
\mathrm{MSE}(s, \hat{s}) = \frac{1}{N} \sum_{i=1}^N \bigl(s(i) - \hat{s}(i)\bigr)^2
\label{eq:mse_definition}
\end{equation}

For clean signals, the normalized PSNR is defined as:
\begin{equation}
y_{\text{clean}} = 1 
\label{eq:y_clean}
\end{equation}

\subsection{Cooperative Learning Process}
\label{subsection:cooperative_learning_process}
Training of CoRe-Net involves a cooperative optimization strategy. The AR and MR are trained with Back Propagation, and their parameters are updated in an iterative manner to ensure that the AR improves its restoration capabilities while the MR refines its evaluation accuracy simultaneously. 

\subsubsection{AR Optimization}  

The AR is optimized to minimize a total restoration loss \( \mathrm{L}_{\text{A}} \), defined as:
\begin{equation}
\mathrm{L}_{\text{A}} = \epsilon \cdot \mathrm{L}_{\text{fid}} + \beta \cdot \mathrm{L}_{\text{time}} + \phi \cdot \mathrm{L}_{\text{freq}}
\label{eq:apprentice_loss}
\end{equation}
where \(\mathrm{L}_{\text{fid}}\), \(\mathrm{L}_{\text{time}}\) and \(\mathrm{L}_{\text{freq}}\) represent fidelity, time-domain, and frequency-domain losses, respectively, and \( \epsilon, \beta, \phi > 0 \) are weights controlling their relative contributions.

The restoration fidelity loss \(\mathrm{L}_{\text{fid}}\) ensures that the AR generates signals \( \hat{s} \)  that closely align with the quality of the clean signals, as predicted by the MR:

\begin{equation}
\mathrm{L}_{\text{fid}} = \mathrm{MSE}(\mathrm{M}(r, \hat{s}), 1) 
\label{eq:fidelity_loss}
\end{equation}

The time-domain loss \(\mathrm{L}_{\text{time}}\) evaluates the restoration quality in the time domain, penalizing deviations from the clean reference \( s \):
\begin{equation}
\mathrm{L}_{\text{time}} = -\mathrm{PSNR}(\hat{s}, s) 
\label{eq:time_domain_loss}
\end{equation}

The frequency-domain loss \( L_{\text{freq}} \) evaluates the
restoration quality in the frequency domain by comparing the
spectrograms of the restored and clean signals

\begin{equation}
\mathrm{L}_{\text{freq}} = -\mathrm{PSNR}(S(\hat{s}), S(s)) 
\label{eq:frequency_domain_loss}
\end{equation}

where \( S(\cdot) \) denotes the spectrogram as calculated in \cite{zahid2024brsr}, promoting correct spectral content.

During the AR update, the MR parameters remain fixed, and gradients from \( L_A \) flow through \( A(\cdot) \), improving its ability to restore radar signals that closely approximate the clean signal \( s \).

\subsubsection{MR Optimization}

Following the AR update, the MR is optimized to refine its predictions of normalized PSNR. The MR minimizes the loss:
\begin{equation}
\mathrm{L}_{\text{M}} = \frac{1}{2} \left( \mathrm{MSE}(\mathrm{M}(r, s), y_{\text{clean}}) + \mathrm{MSE}(\mathrm{M}(r, \hat{s}), y_{\text{res}}) \right)
\label{eq:master_loss}
\end{equation}

where \( \mathrm{M}(r, s) \) predicts a PSNR value near \(1\) for the clean signal \( s \), and \( \mathrm{M}(r, \hat{s}) \) predicts \( y_{\text{restored}} \), the normalized PSNR of \( \hat{s} \) relative to \( s \). By minimizing \( \mathrm{L}_\text{M} \), the MR becomes a reliable quality estimator, indirectly guiding the AR's restoration process through \( \mathrm{L}_\text{fid} \). By leveraging the MR’s feedback, the AR no longer depends on binary discrimination but instead uses a regression-based quality measure. This cooperative setup enables stability and a better learning capability, which in turn leads to higher-quality restorations.

The cooperative feedback between the AR and MR is established through \( \mathrm{L}_\text{fid} \), where the MR's evaluation of \( \hat{s} \) directly influences the AR's updates. This interaction enables the AR to iteratively refine its restorations based on the MR’s guidance, ensuring that the generated outputs achieve high-quality restoration. By employing task-specific regression instead of adversarial discrimination, CoRe-Net avoids challenges such as mode collapse and achieves stable convergence. As the MR becomes a more reliable quality evaluator, the AR progressively enhances its ability to restore radar signals effectively. The training procedure, detailed in Algorithm~\ref{algo:corenet_training}, alternates between optimizing the parameters of AR and MR, fostering a cooperative mechanism that simultaneously improves the AR's restoration performance in the end.

\begin{algorithm}[t!]
\caption{Training CoRe-Net Framework}
\label{algo:corenet_training}
\small
\begin{algorithmic}[1]
\State \textbf{Input:} Training samples $\{(r_i, s_i)\}_{i=1}^N$, hyperparameters $\epsilon$, $\beta$, $\phi$, learning rates $\eta_A$, $\eta_M$, maximum iterations $\text{maxIter}$.
\State \textbf{Output:} Trained weights for AR and MR: $\Theta_A$, $\Theta_M$.

\State Initialize trainable parameters $\Theta_A$ and $\Theta_M$.
\State Normalize all training samples $r_i$ and $s_i$ to the range $[-1, 1]$.

\For{$t = 1$ to $\text{maxIter}$}
    \State Sample a mini-batch $(r, s)$ from the training dataset.

    \Comment{\textit{Step 1: Update AR}}
    \State Generate restored signal $\hat{s} = \mathrm{A}(r)$.
    \State Compute the fidelity loss \( \mathrm{L}_{\text{fid}} \) as in Eq.~\eqref{eq:fidelity_loss}.
    \State Compute the time-domain PSNR loss \( \mathrm{L}_{\text{time}} \) as in Eq.~\eqref{eq:time_domain_loss}.
    \State Compute the frequency-domain PSNR loss \( \mathrm{L}_{\text{freq}} \) as in Eq.~\eqref{eq:frequency_domain_loss}.
    \State Compute the total Apprentice loss \( \mathrm{L}_{\text{A}} \) as in Eq.~\eqref{eq:apprentice_loss}.
    \State Update AR weights $\Theta_A$ using Adam optimizer to minimize \( L_A \).

    \Comment{\textit{Step 2: Update MR}}
    \State Predict normalized PSNR for clean and restored signals:
    \[
    M_{\text{clean}} = \mathrm{M}(r, s), \quad M_{\text{restored}} = \mathrm{M}(r, \hat{s}).
    \]
    \State Compute the target label \( y_{\text{res}} \) for the restored signal as in Eq.~\eqref{eq:y_res}.
    \State Compute the Master loss \( \mathrm{L}_{\text{M}} \) as in Eq.~\eqref{eq:master_loss}.
    \State Update MR weights $\Theta_M$ using Adam optimizer to minimize \( L_M \).
\EndFor

\State \textbf{return} $\Theta_A$, $\Theta_M$.
\end{algorithmic}
\end{algorithm}

\subsection{Progressive Transfer Learning (PTL)}

\begin{figure}[t!]
\centering
\includegraphics[width=\linewidth]{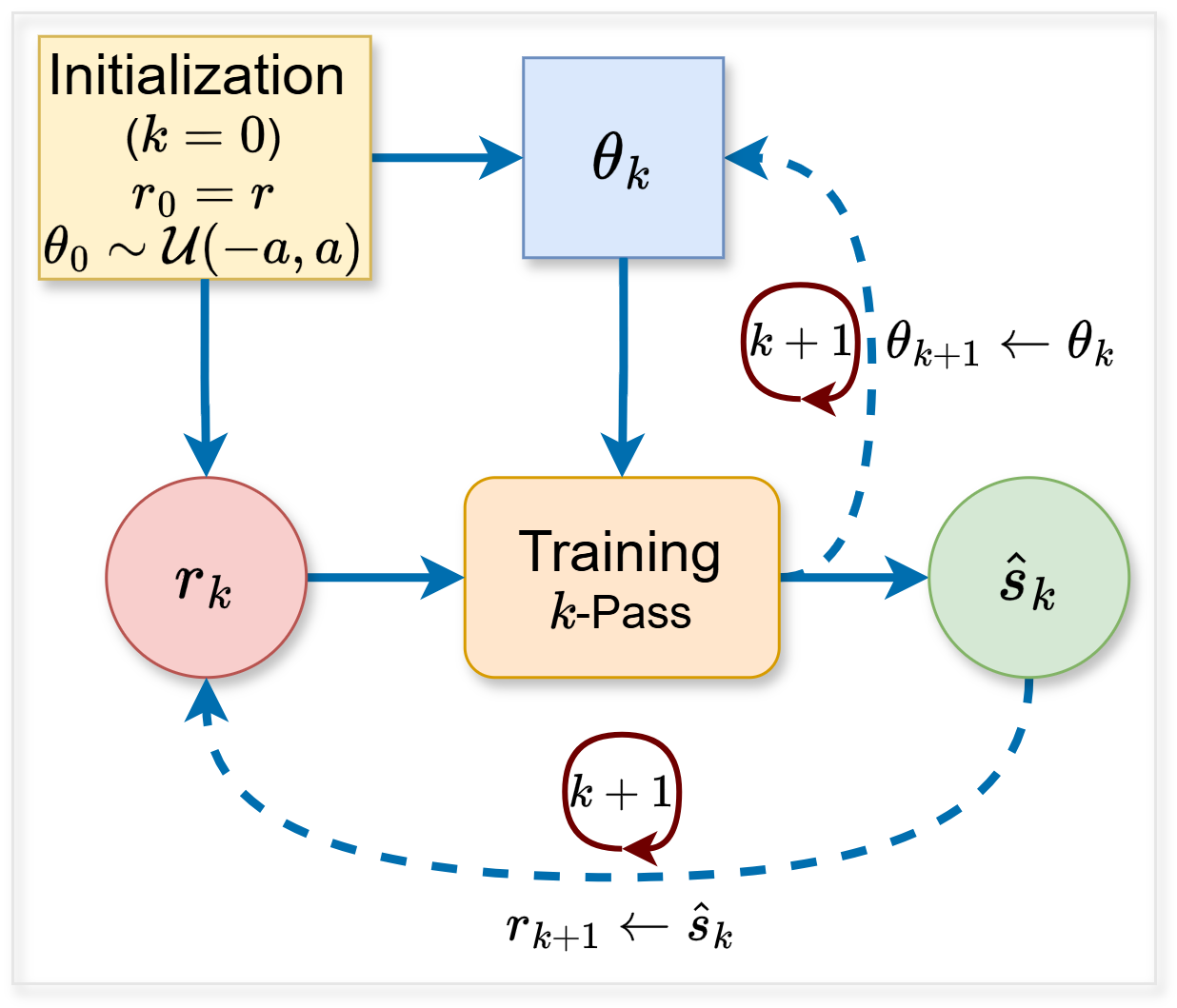}
 \caption{Progressive Transfer Learning (PTL) strategy for CoRe-Net. Each training pass (\(k\)) starts with a distorted signal (\(r_k\)) and produces a restored signal (\(\hat{s}_k\)), which becomes the input for the next pass (\(r_{k+1}\)). Parameters (\(\theta_k\)) are initialized using Xavier uniform initialization for \(k=0\) and updated iteratively using the best-performing weights from the previous pass based on the validation SNR.}
\label{fig:ptl_workflow}
\end{figure}

To further boost the restoration performance, multiple training passes can be executed to create a cascaded series of ARs where each pass uses the AR's outputs from the previous pass as the inputs so that such a consecutive restoration can address the artifacts that still persist in previous pass(es) as well as the new artifacts possibly occurred as a result of earlier restorations. After the first pass where the network parameters of both AR and MR are randomly initialized, the Progressive Transfer Learning (PTL) aims to pass the learned "restoration knowledge" in each pass to the next pass. Therefore, PTL  ensures that the training of both AR and MR at each pass \(k\) starts from the parameters of the previous pass. So instead of a random initialization at pass \(k+1\), both AR and MR networks are initialized with the parameters of the best corresponding network parameters converged in the pass \(k\). Naturally, in the current PTL implementation, in a particular training pass we select the CoRe-Net model that yields the best restoration performance in the validation set as the best network to be inherited during the next training pass and so on. In this way, the PTL enables the model to start its training in the best possible (learned in advance) point in the parameter space, and such an internal transfer learning mechanism progressively improves the overall learning and restoration performances. This PTL strategy is illustrated in Fig.~\ref{fig:ptl_workflow}.

At the first pass (\(k=0\)), the input signal is defined as:
\begin{equation}
r_0 = r
\end{equation}
where \(r\) represents the original distorted radar signal. The parameters of CoRe-Net, comprising both the AR and the MR, are initialized using Xavier uniform initialization \cite{glorot2010understanding}.

During each training pass \(k\), the CoRe-Net model parameters are optimized by minimizing the CoRe-Net loss functions in a cooperative manner, as described in ~\ref{subsection:corenet}. After a training pass is complete, the AR restores the distorted signal \(r_k\) to produce the corresponding output \(\hat{s}_k\):
\begin{equation}
\hat{s}_k = \mathrm{A}_{\theta_k}(r_k)
\end{equation}
where \(A_{\theta_k}\) denotes the AR parameterized by \(\theta_k\). 

Then the restored signal \(\hat{s}_k\) is fed back into the next training pass as the input:
\begin{equation}
r_{k+1} \leftarrow \hat{s}_k
\end{equation}
Similarly, the network parameters of both AR and MR for the next training pass are initialized using the best-performing configuration parameters converged in the current pass, selected based on the validation SNR:
\begin{equation}
\theta_{k+1} \leftarrow \underset{\theta \in \Theta_k}{\arg\max} \; \mathrm{SNR}_{\text{val}}^{k}
\end{equation}

This internal transfer learning mechanism ensures that both inputs and parameters are progressively refined, enabling CoRe-Net to handle increasingly challenging distortions in multiple learning sessions. The workflow in Fig.~\ref{fig:ptl_workflow} highlights this process, showcasing how the input signal (\(r_k\)) and parameters (\(\theta_k\)) are iteratively updated to improve restoration quality.  As a result, PTL can enhance the model’s generalization capability for blind restoration across diverse artifact scenarios.

\begin{figure*}[t!]
    \centering
    \includegraphics[width=1\textwidth,keepaspectratio]{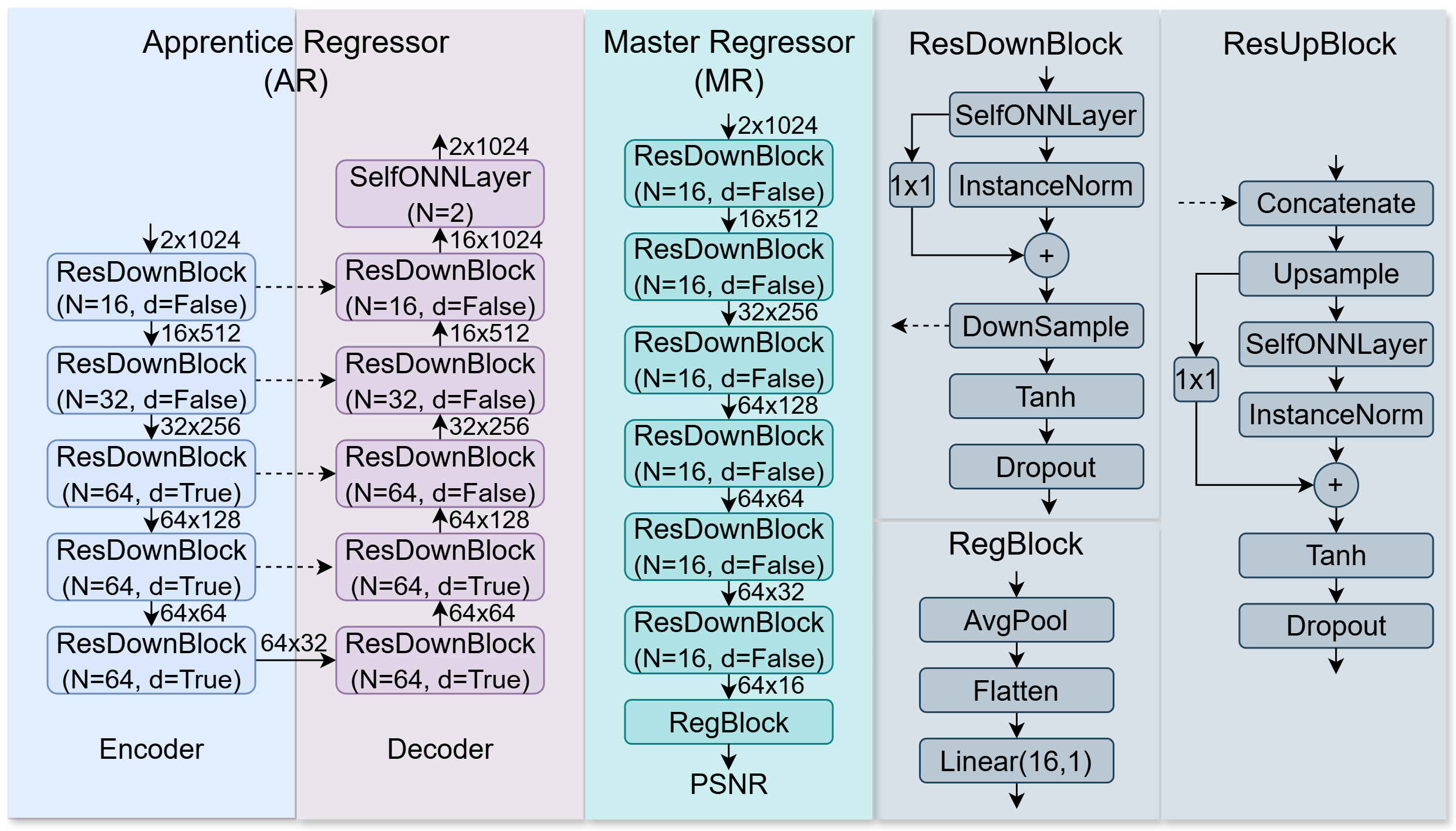}
    \caption{Architecture of CoRe-Net showcasing the Apprentice Regressor (AR) and the Master Regressor (MR). The AR, on the left, is divided into encoder and decoder parts. The MR is shown in the center. Both components share building blocks such as ResDownBlock, ResUpBlock, and the Regression Block (right). \(N\) represents the number of filters per block, while \(d\) indicates the use of dropout. Skip connections are depicted with dotted lines, and feature dimensions are labeled at each stage. The Regression Block in the MR predicts PSNR (scaled between 0 and 1). The color coding of AR and MR matches that of the methodology figure for consistency.}
    \label{fig:architecture}
\end{figure*}

\subsection{CoRe-Net Architectures}
The overall architecture of Core-Net is illustrated in Fig.~\ref{fig:architecture}. The AR follows an encoder-decoder architecture that transforms corrupted radar signals $\hat{x}$ into restored outputs $\tilde{x}$. The encoder consists of five Residual Downsampling Blocks (ResDownBlocks) integrating 1D Self-Organized Operational Neural Network (Self-ONN) layers, instance normalization, and $\tanh$ activations. These layers enhance non-linear feature learning and stabilize the training process. Correspondingly, the decoder employs five Residual Upsampling Blocks (ResUpBlocks) to reconstruct the restored signal. Skip connections between the encoder and decoder layers preserve low-level features and ensure accurate reconstruction of fine-grained signal details. 

The MR is designed to evaluate the quality of restored radar signals by predicting their PSNR. The MR follows a residual architecture comprising six Residual Downsampling Blocks (ResDownBlocks) and a final regression block, which includes adaptive average pooling and a fully connected layer to output normalized PSNR values within the range $[0, 1]$.

The MR incorporates a distortion-aware mechanism to enhance its evaluation capability by accepting pairs of signals as inputs. Its objective is to regress (estimate) the PSNR of the restored $\tilde{x}$ and clean $x$ signals and use this to assist the AR continuously during training. To make this estimation more accurate, the corrupted (input) signal $\hat{x}$ is also fed into MR so that it can quantify the quality improvement better by comparing the restored/clean signal with the original (corrupted) version. This mechanism also improves the training stability of MR.

Both the AR and MR process radar signal segments of 1024 samples, divided into two channels representing the in-phase (I) and quadrature (Q) components. The 1D Self-ONN layers used in ResDownBlocks and ResUpBlocks employ a kernel size of $1 \times 3$, with a dropout rate of $d = 0.25$ to mitigate overfitting.

\section{BRSR Dataset Overview}
\label{sec:dataset}

Real-world radar signals often suffer from various artifacts and this makes target detection or recognition quite challenging and sometimes infeasible. These artifacts arise from environmental factors, system imperfections, and external disturbances. This section first summarizes the common artifacts encountered in Real-World radar signal processing and the creation of the BRSR benchmark dataset used for evaluation.

\subsection{Artifacts in Real-World Radar Signals}
Radar signals in Real-World scenarios are often corrupted by a random blend of artifacts each of which can also have a random scale. Accordingly, this is modeled in this study as a composite signal distortion using the most common artifacts with randomized weights~\cite{zahid2024brsr}. The composite distortion can be expressed as:
\begin{equation}
r(t) = s(t) + w_1 \cdot n(t) + w_2 \cdot s(t-\tau) + w_3 \cdot i(t)
\label{eq:noise_composite}
\end{equation}
where \(n(t)\) represents Additive White Gaussian Noise (AWGN) with zero mean and variance \(\sigma^2\), \(s(t-\tau)\) denotes an echo signal delayed by \(\tau\) and scaled by an attenuation factor \(w_2\), and \(i(t)\) is an interfering signal. The weights \(w_1\), \(w_2\), and \(w_3\) are assigned randomly to control the contribution of AWGN, echo, and interference, respectively, along with the possibility of setting one or more weights to zero. This approach ensures a diverse and realistic representation of radar signal corruption, facilitating a challenging testbed for blind restoration methods.

\subsection{BRSR Benchmark Dataset}
The BRSR dataset~\cite{zahid2024brsr} provides a benchmark platform for testing blind radar signal restoration algorithms. It includes paired clean and corrupted radar signals generated by accumulating common artifacts with randomized weights on the clean signal. Therefore, it represents an actual received radar signal corrupted with random set of artifacts. Key features of the BRSR dataset can be summarized below.

\begin{itemize}
    \item Signal Representation Each radar signal is represented by a two-channel (real and imaginary) complex-valued signal, with a segment length of 1024 samples.
    \item Artifact Combinations: The dataset includes signals corrupted by individual artifacts, dual combinations, and all three artifacts, providing a comprehensive spectrum of distortion patterns.
    \item SNR Range: Signals are corrupted to achieve an SNR range of \([-14 , 10] \, \text{dB}\), with artifact severities adjusted using random weights and normalized to maintain the overall SNR.
    \item Dataset Size: The dataset comprises a total of 85,800 paired samples. Of these, 62,400 samples are allocated for training (80\% ) and validation (20\%). Furthermore, 150 signals are generated at each SNR level, resulting in 23,400 test signals. The split remains the same as in ~\cite{zahid2024brsr} to ensure a fair comparative evaluation.

\end{itemize}

This dataset enables comprehensive evaluation across diverse SNR levels and artifact combinations, ensuring an effective assessment framework. For artifact-specific details, including random parameter generation (e.g., echo delays, interference selection, and artifact weights), readers are referred to~\cite{zahid2024brsr}.

\subsection{Data Preprocessing and Normalization}
Before training, all radar signals are normalized independently for each channel (in-phase and quadrature) to the range \([-1, 1]\), ensuring consistency across varying signal amplitudes. The normalization process is defined as:
\begin{equation}
X_N^s(i) = 2 \cdot \frac{X^s(i) - X_{\mathrm{min}}^s}{X_{\mathrm{max}}^s - X_{\mathrm{min}}^s} - 1
\label{eq:normalization}
\end{equation}
where \(X^s(i)\) represents the signal value at the \(i\)-th sample in segment \(s\), \(X_{\mathrm{min}}^s\) and \(X_{\mathrm{max}}^s\) are the minimum and maximum values of the segment, and \(X_N^s(i)\) is the normalized value.

This normalization not only ensures amplitude consistency by standardizing signal amplitudes, eliminating variations due to differing scales, but is also critical for the stability and effectiveness of the Self-ONN layers employed in the CoRe-Net framework.

Normalization further improves generalization by removing amplitude-related biases, enabling the model to perform robustly across diverse SNR levels and artifact severities. Moreover, consistent input scales contribute to stable training, preventing the model from being influenced by outlier amplitude values. This preprocessing step is particularly important for datasets like BRSR, which encompass a wide range of SNR levels and include various artifact combinations, ensuring the robustness and adaptability of the proposed CoRe-Net framework.

\section{EXPERIMENTAL RESULTS}
\label{sec:results}
This section presents a comprehensive evaluation of the proposed CoRe-Net framework. First, we outline the experimental setup and the evaluation metrics employed. Subsequently, we perform a detailed comparative analysis of CoRe-Net against the prior state-of-the-art model, BRSR-OpGAN, over the extended BRSR dataset. The analysis includes both quantitative and qualitative assessments, highlighting the restoration performance across various radar signal modulation types and training passes. Finally, we examine the computational efficiency of CoRe-Net, emphasizing its practicality for real-time radar signal restoration.

\subsection{Experimental Setup}
The training procedure for CoRe-Net is configured to run for a maximum of 1000 back-propagation (BP) epochs with a batch size of 64. The Adam optimizer is employed, with the initial learning rates for both the AR and MR set at \(5 \cdot 10^{-3}\) and cosine annealing learning rate scheduler (CosineAnnealingLR) to ensure smooth convergence.  The cosine annealing strategy, as described in \cite{loshchilov2016sgdr}, provides a dynamic learning rate adjustment, improving stability and performance during training. The \(t_{\text{max}}\) parameter for the scheduler is set to 100 iterations. The loss weights in \eqref{eq:apprentice_loss} are optimized using the Optuna library \cite{akiba2019optuna} with 1000 trials. These weights are selected to maximize the validation SNR, with searches conducted within the ranges \( \epsilon, \beta \in [1, 10] \) and \( \phi \in [1, 5] \), ensuring optimal contributions from loss components.

A comprehensive set of experiments is conducted to find the best parameter configuration for CoRe-Net. Specifically, we varied the target PSNR (\(30 \, \text{dB}\) or \(40 \, \text{dB}\)), along with different values for \(\varepsilon\), \(\beta\), and \(\phi\). We also investigated the effect of using either a cosine annealing or constant learning rate scheduler and compared MSE (L2) versus L1 loss as the MR loss function. Finally, we evaluated both single- and dual-channel inputs for the MR. Based on the validation SNR, the best performance was obtained when the target PSNR was set to \( 40 \, \text{dB} \) with \(\varepsilon=1\), \(\beta=10\), and \(\phi=1\); the MR loss function was MSE; the learning rate scheduler used cosine annealing; and the MR received a dual-channel input. Consequently, these parameters were used for subsequent CoRe-Net experiments.

To ensure a fair comparison between the adversarial training of OpGANs and the proposed cooperative training of CoRe-Nets, we used same configurations and parameters of the network architectures of BRSR-OpGAN and CoRe-Net. The order of the Taylor polynomials in all Self-ONN layers, Q, is set as 3. The implementation of 1D Self-ONN layers in CoRe-Net is facilitated by the FastONN library \cite{malik2020fastonn}, built on Python and PyTorch. This consistent setup guaranteed the reliability and fair comparability of all results.

Signal to Noise Ratio (SNR) is used as the common metric to quantify the restoration performance that can be expressed as follows:

\begin{equation}
\mathrm{SNR} = 10 \log_{10} \left( \frac{\sum_{i=1}^N s(t_i)^2}{\sum_{i=1}^N [s(t_i) - \hat{s}(t_i)]^2} \right)
\end{equation}
where \(s(t)\) and \(\hat{s}(t)\) are the clean and restored signals, respectively, and \(N\) is the number of samples.

\subsection{ Comparative Evaluations }
As presented in Table~\ref{tab:comparison_methods},  under fair conditions and setup, the proposed CoRe-Net model has surpassed all the prior state-of-the-art models including the most recent GAN model, BRSR-OpGAN~\cite{zahid2024brsr}  with a significant performance gap over the extended BRSR dataset.   

\begin{table}[t!]
\centering
\caption{Restoration performances of the CoRe-Net and prior state-of-the-art methods.}
\label{tab:comparison_methods}
\resizebox{\linewidth}{!}{
\begin{tabular}{lc}
    \toprule
    \textbf{Restoration Algorithms} & \textbf{Average SNR (dB)} \\
    \midrule
    \cellcolor[gray]{.93}Corrupted Signal & \cellcolor[gray]{.93}-1.94 \\
    VMD-LMD-WT \cite{jiang2024multilayer} & 3.36 \\
    \cellcolor[gray]{.93}CNN-GAN & \cellcolor[gray]{.93}9.04 \\
    BRSR-OpGAN \cite{zahid2024brsr} & 10.30 \\
    \cellcolor[gray]{.93}CoRe-Net & \cellcolor[gray]{.93}\textbf{11.37} \\
    \bottomrule
\end{tabular}
}
\end{table}

Starting with the decomposition-based method VMD-LMD-WT~\cite{jiang2024multilayer}, the results show marginal improvements over the corrupted reference signal, achieving an average SNR of \( 3.36 \, \text{dB} \). This reflects the limitations of traditional signal decomposition approaches in performing blind restoration over such complex, multi-faceted signal corruptions.

Significant improvements are observed with the first and traditional deep learning model, the CNN-GAN, leveraging both time and frequency domain losses, achieves an average SNR of \( 9.04 \, \text{dB} \), demonstrating the importance of incorporating domain-specific insights into restoration tasks. This model serves as a strong baseline for GAN-based restoration techniques.

Sharing the same network architecture as CNN-GAN, the most recent GAN model based on 1D Self-ONNs, the BRSR-OpGAN \cite{zahid2024brsr} outperforms the CNN-GAN by more than 1dB in SNR. It achieves an SNR level of \( 10.30 \, \text{dB} \), setting the best performance so far achieved for blind radar signal restoration. These results highlight the effectiveness of leveraging Self-ONN layers along with the dual domain loss.

Finally, the proposed CoRe-Net model with the same number of parameters and complexity achieves an SNR of \( 11.37 \, \text{dB} \), surpassing BRSR-OpGAN by more than \( 1 \, \text{dB} \). Such a crucial performance gain underscores the superiority of CoRe-Net's \textit{cooperative learning paradigm}, which replaces adversarial training with task-specific collaboration between the two regressors simultaneously trained. The CoRe-Net achieves such a blind restoration performance in a single training pass; thus, showcasing its efficiency and robustness in handling diverse distortion scenarios.

\begin{figure*}[t!]
\centering
\includegraphics[width=\linewidth]{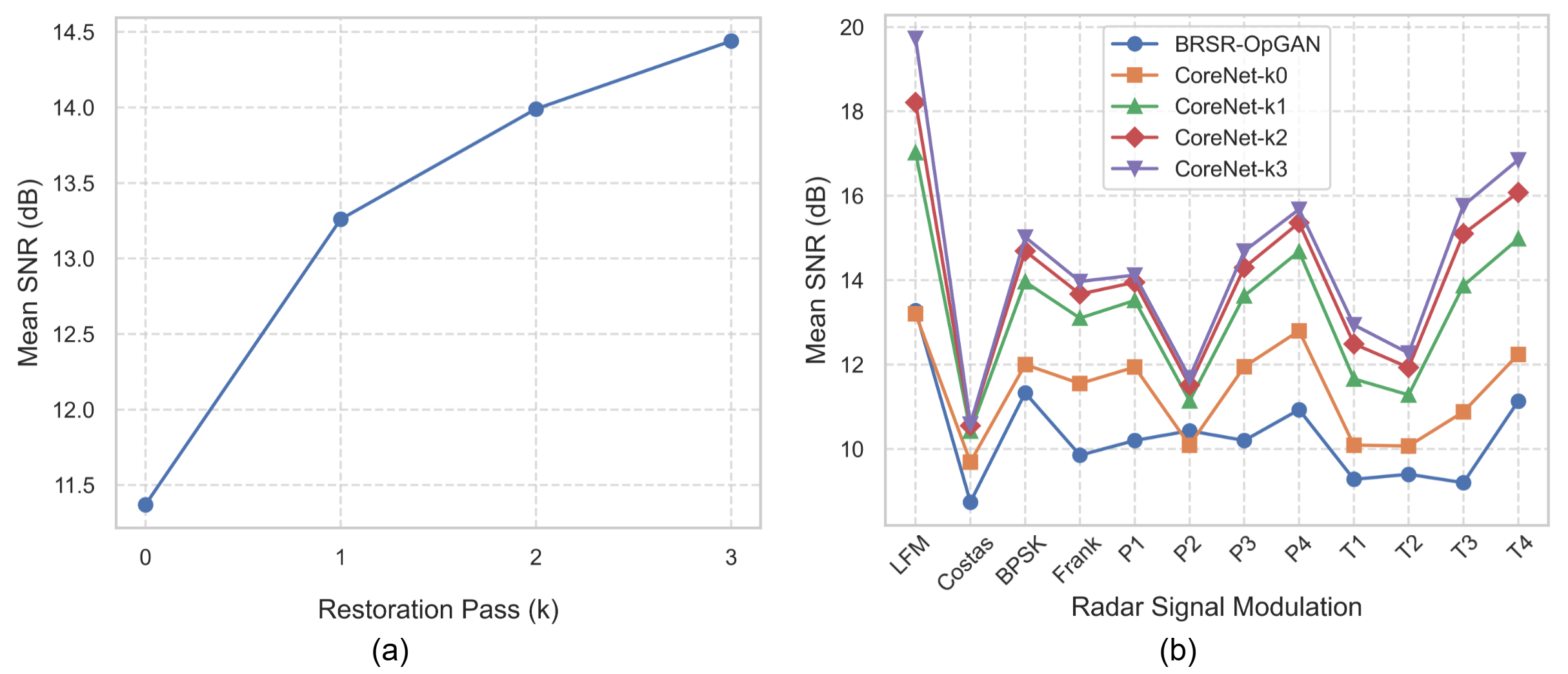}
\caption{Restoration performance of CoRe-Net across training passes and modulation types. Mean SNR across four restoration passes with PTL (a), demonstrating significant gains in restoration quality. Mean SNR for radar signal modulation type (b), comparing the performance of CoRe-Net across multiple passes and BRSR-OpGAN.}
\label{fig:snr_ptl_modulation}
\end{figure*}

\subsection{ Multi-pass Restoration Results  }
Fig.~\ref{fig:snr_ptl_modulation}(a) plots the restoration performance of CoRe-Nets trained with 4 training passes with Progressive Transfer Learning (PTL) on the BRSR dataset. It is clear that PTL leads to substantial restoration gains (\(\sim 1.9 \, \text{dB}\) in SNR) by the CoRe-Net trained in the first training pass after the initial training (i.e., SNR of \(11.37 \, \text{dB}\) at \(k=0\)). The following two training passes with PTL lead to more than \(3 \, \text{dB}\) SNR gain after the initial training, which is crucial. The overall mean restoration SNR now exceeds \(14 \, \text{dB}\) overall. The results demonstrate consistent improvements in the restoration quality highlighting PTL's crucial role in the multi-pass training that leads to a multi-pass restoration with an unprecedented performance.

\subsection{Restoration Performance for Each Modulation Type}

Fig.~\ref{fig:snr_ptl_modulation}(b) illustrates the restoration performances in terms of SNR improvements achieved by CoRe-Net and BRSR-OpGAN across various radar signal modulation types. The x-axis represents different modulation schemes, such as LFM, Costas, BPSK, etc., while the y-axis denotes the mean SNR improvement (in dB) compared to the mean SNR level of the distorted signals.

The results reveal a clear trend of increasing and mostly substantial SNR improvements as higher the restoration passes by CoRe-Nets  (k0 to k3). CoReNet-k3 consistently achieves the highest SNR improvement across all modulation types as expected. Notably, the improvement is particularly significant for modulation types such as LFM and T4.

CoRe-Net outperforms BRSR-OpGAN in all modulation types except the modulation type \( P2 \), where BRSR-OpGAN achieves slightly higher SNR improvement. However, the second restoration pass by CoRe-Net again outperforms BRSR-OpGAN. In contrast, for most modulation types, BRSR-OpGAN demonstrates limited SNR improvement, particularly for complex schemes. While it provides some enhancement for simpler modulation types, such as BPSK and Frank, its performance plateaus for more challenging cases. This highlights the limitations of traditional GAN-based approaches in handling diverse artifact severities and modulations.

\subsection{Qualitative Evaluation}
\begin{figure*}[!htb]
\centering
\includegraphics[width=\textwidth,keepaspectratio]{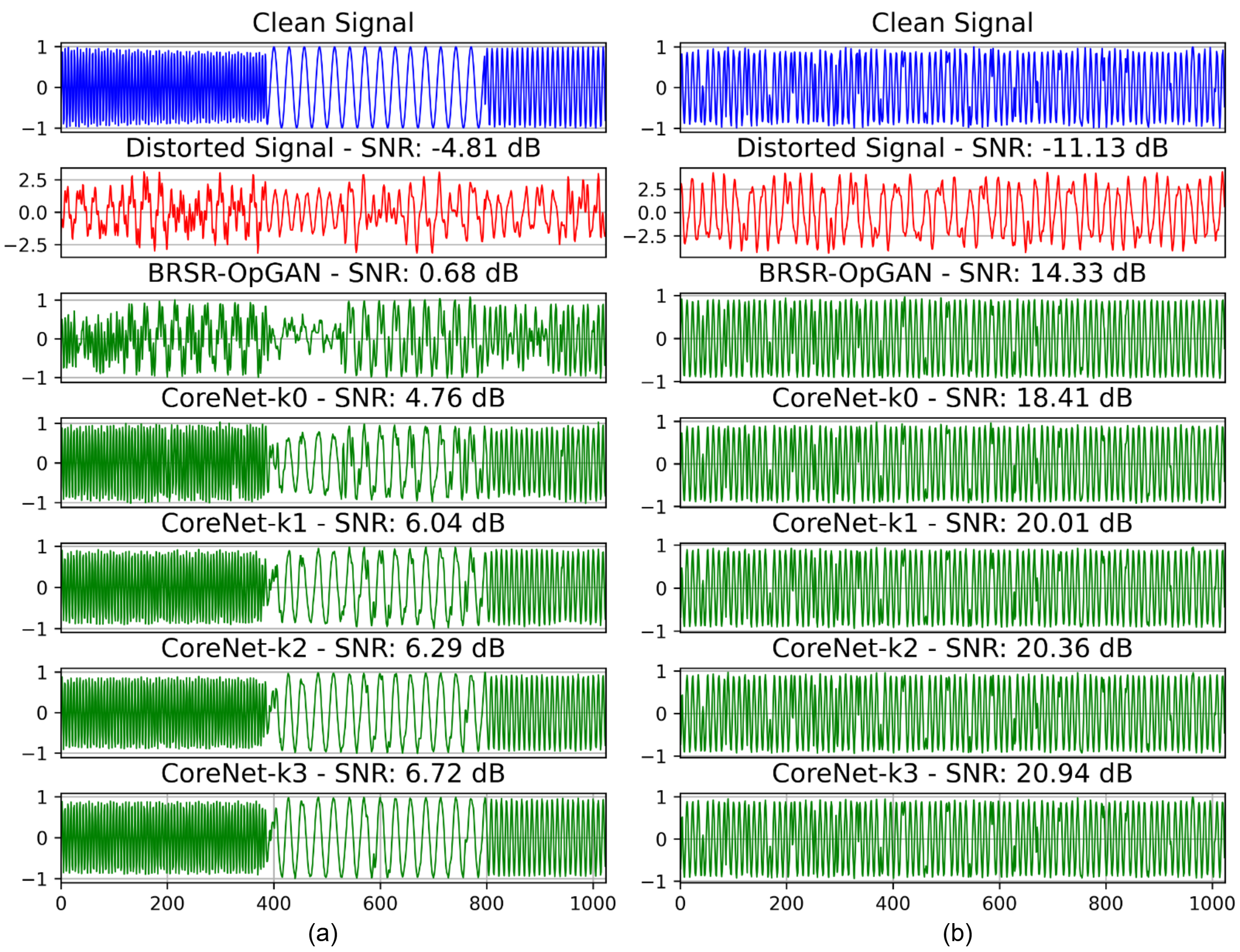}
\caption{Sample radar signals corrupted with random blend artifacts and the corresponding restored signals.}
\label{fig:qualitative_1}
\end{figure*}

Fig.~\ref{fig:qualitative_1} presents a detailed visual comparison of radar signal restoration performance for CoRe-Net and BRSR-OpGAN. In each subfigure, the top plot illustrates the original transmitted radar signal \(s(t)\), which serves as the clean (target) signal. The second plot displays the received signal \(r(t)\), which is severely corrupted by artifacts such as noise, echo, and interference. The subsequent plots depict the restored signals \(\hat{s}(t)\) produced by BRSR-OpGAN and CoRe-Nets with multi-pass training by PTL, labeled as CoReNet-k0 through CoReNet-k3.

The figure demonstrates the capability of CoRe-Net to progressively refine the restoration quality through its restoration passes. Even in the initial pass (CoReNet-k0), it generates a highly superior restoration compared to BRSR-GAN. However, minor distortions still remain, particularly in regions with abrupt signal variations. They are addressed by the second and third pass restorations and finally, at the fourth pass (CoReNet-k3), the difference between the restored and clean signal entirely diminishes. Such a multipass blind restoration approach can minimize the diverse set of artifacts with random severities and types, effectively addressing challenges posed by high-frequency variations and complex radar signal structures.

\subsection{Computational Complexity Analysis}
During the inference time, the AR of CoRe-Net and the Generator of BRSR-OpGAN are used for restoration. As discussed earlier, identical models with 1D Self-ONN layers are used in both networks. Therefore, one can expect the same number of parameters, network depth, and inference times for a single-pass restoration. To validate this, we calculate the number of parameters (PARs) and the inference time required for a batch size of 64 for each network configuration. The detailed formulations for calculating PARs in Self-Organizing Neural Networks (Self-ONNs) are provided in \cite{malik2020fastonn}. All experiments are conducted on a system with a 2.4 GHz Intel Core i7 processor, 32 GB of RAM, and an NVIDIA GeForce RTX A1000 GPU with 6 GB of VRAM. While training and testing were performed on GPU cores, inference times are also measured using a single CPU to assess lightweight deployment scenarios.

As presented in Table~\ref{tab:computation}, the AR of CoRe-Net and the Generator (G) of BRSR-OpGAN yield almost identical computational complexities in terms of parameters and inference times. However, multi-pass restoration by CoRe-Nets will increase the complexity proportional to the number of passes. But considering the significantly low computational complexity levels both in terms of parameters (memory) and inference times, performing a few restoration passes by CoRe-Nets will still permit the overall radar signal restoration in real time with a lightweight implementation. Considering the crucial gains in the overall restoration performance, multi-pass restoration can naturally be a favorable option over a single pass in practice.

\begin{table}
\centering
\caption{Computational Complexity of CoRe-Net and BRSR-OpGAN. Parameters (PARs) are measured in thousands (K), and inference times are reported in milliseconds (ms). Inference times are provided for both CPU and GPU execution.}
\label{tab:computation}
\begin{tabular}{lcccccc}
    \toprule
    \textbf{Model} & \multicolumn{3}{c}{\textbf{Parameters (PARs) (K)}} & \multicolumn{2}{c}{\textbf{Inference Time (ms)}} \\
    \cmidrule(lr){2-4} \cmidrule(lr){5-6}
    & \textbf{G/AR} & \textbf{D/MR} & \textbf{Total} & \textbf{CPU} & \textbf{GPU} \\
    \midrule
    Core-Net & 275.52 & 82.61 & 358.13 & 0.710 & 0.039 \\
    BRSR-OpGAN & 275.52 & 82.61 & 358.13 & 0.711 & 0.041 \\
    \bottomrule
\end{tabular}
\end{table}

\section{CONCLUSION}
\label{sec:conclusion}
The primary challenge in restoring real-world radar signals is the impairment caused by numerous artifacts, including echo, sensor noise, jamming, and atmospheric disturbances. These artifacts differ significantly in type, intensity, and duration, resulting in highly corrupted signals and this makes a robust and effective restoration challenging. Despite numerous prior research in radar signal denoising and interference mitigation, no comprehensive solution has been developed to address these diverse challenges.

This study introduces Co-operational Regressor Network (CoRe-Net), a pioneer model designed for blind radar signal restoration. By leveraging a collaborative learning paradigm between the two regressors, Apprentice Regressor (AR) and Master Regressor (MR), CoRe-Net addresses the limitations of traditional adversarial models. This cooperative approach ensures stability and adaptability, enabling a superior blind signal restoration without assuming any artifact type and severity in advance. Additionally, performing multi-pass restoration with multiple CoRe-Nets, each of which is trained by the proposed Progressive Transfer Learning (PTL), boosts the restoration performance by more than \(3 \, \text{dB}\) in SNR, achieving significant performance gains with each pass.

An extensive set of comparative evaluations on the benchmark BRSR dataset, CoRe-Net demonstrated state-of-the-art restoration performance, surpassing all prior methods by more than \(1 \, \text{dB}\) in SNR. By employing compact and computationally efficient Self-Organized Operational Neural Network (Self-ONN) layers, CoRe-Net achieves this performance without increasing computational complexity, making it suitable for real-time applications on resource-constrained platforms. The performance gap exceeds \(3 \, \text{dB}\) in SNR when 3-pass restoration is performed by CoRe-Nets trained consecutively with PTL. Though it increases the inference time and complexity by three times, the lightweight CoRe-Net architecture still allows a real-time implementation. 

Future work will focus on extending the CoRe-Net model to incorporate classification into the restoration process, aiming to improve the regression further to maximize the classification performance (and vice versa) on the same network. Additionally, exploring a hybrid learning paradigm that combines cooperative learning, as demonstrated in CoRe-Net, with competitive learning, as utilized in GANs, presents a compelling direction. This approach mirrors how humans learn through collaboration and competition, leveraging their complementary strengths to achieve balanced and robust learning. Further efforts will also investigate expanding the applicability of CoRe-Net to address more complex artifact scenarios and diverse signal types, while adaptive PTL strategies will be explored to optimize the trade-off between performance improvement and computational cost. These advancements further improve CoRe-Net's position as a leading framework for other blind signal restoration tasks and even its applicability to other regression tasks.

\section*{Acknowledgments}
This work was supported by the Business Finland project AMaLIA and NSF Center for Big Learning (CBL).

\bibliographystyle{IEEEtran}
\bibliography{refs}

\begin{thebibliography}{10}
\providecommand{\url}[1]{#1}
\csname url@samestyle\endcsname
\providecommand{\newblock}{\relax}
\providecommand{\bibinfo}[2]{#2}
\providecommand{\BIBentrySTDinterwordspacing}{\spaceskip=0pt\relax}
\providecommand{\BIBentryALTinterwordstretchfactor}{4}
\providecommand{\BIBentryALTinterwordspacing}{\spaceskip=\fontdimen2\font plus
\BIBentryALTinterwordstretchfactor\fontdimen3\font minus \fontdimen4\font\relax}
\providecommand{\BIBforeignlanguage}[2]{{%
\expandafter\ifx\csname l@#1\endcsname\relax
\typeout{** WARNING: IEEEtran.bst: No hyphenation pattern has been}%
\typeout{** loaded for the language `#1'. Using the pattern for}%
\typeout{** the default language instead.}%
\else
\language=\csname l@#1\endcsname
\fi
#2}}
\providecommand{\BIBdecl}{\relax}
\BIBdecl

\bibitem{de2018introduction}
A.~De~Martino, \emph{Introduction to modern EW systems}.\hskip 1em plus 0.5em minus 0.4em\relax Artech house, 2018.

\bibitem{tsui1986microwave}
J.~B.-Y. Tsui, ``Microwave receivers with electronic warfare applications,'' \emph{(No Title)}, 1986.

\bibitem{pace2009detecting}
P.~E. Pace, \emph{Detecting and classifying low probability of intercept radar}.\hskip 1em plus 0.5em minus 0.4em\relax Artech house, 2009.

\bibitem{zhou2019ensemble}
Z.~Zhou, G.~Huang, and X.~Wang, ``Ensemble convolutional neural networks for automatic fusion recognition of multi-platform radar emitters,'' \emph{ETRI Journal}, vol.~41, no.~6, pp. 750--759, 2019.

\bibitem{kishore2017automatic}
T.~R. Kishore and K.~D. Rao, ``Automatic intrapulse modulation classification of advanced lpi radar waveforms,'' \emph{IEEE Transactions on Aerospace and Electronic Systems}, vol.~53, no.~2, pp. 901--914, 2017.

\bibitem{seddighi2020radar}
Z.~Seddighi, M.~R. Ahmadzadeh, and M.~R. Taban, ``Radar signals classification using energy-time-frequency distribution features,'' \emph{IET Radar, Sonar \& Navigation}, vol.~14, no.~5, pp. 707--715, 2020.

\bibitem{jan2020artificial}
M.~Jan and D.~Pietrow, ``Artificial neural networks in the filtration of radiolocation information,'' in \emph{2020 IEEE 15th International Conference on Advanced Trends in Radioelectronics, Telecommunications and Computer Engineering (TCSET)}.\hskip 1em plus 0.5em minus 0.4em\relax IEEE, 2020, pp. 680--685.

\bibitem{zhou2018automatic}
Z.~Zhou, G.~Huang, H.~Chen, and J.~Gao, ``Automatic radar waveform recognition based on deep convolutional denoising auto-encoders,'' \emph{Circuits, Systems, and Signal Processing}, vol.~37, pp. 4034--4048, 2018.

\bibitem{zhang2017convolutional}
M.~Zhang, M.~Diao, and L.~Guo, ``Convolutional neural networks for automatic cognitive radio waveform recognition,'' \emph{IEEE access}, vol.~5, pp. 11\,074--11\,082, 2017.

\bibitem{jiang2024multilayer}
M.~Jiang, F.~Zhou, L.~Shen, X.~Wang, D.~Quan, and N.~Jin, ``Multilayer decomposition denoising empowered cnn for radar signal modulation recognition,'' \emph{IEEE Access}, 2024.

\bibitem{lee2019mutual}
S.~Lee, J.-Y. Lee, and S.-C. Kim, ``Mutual interference suppression using wavelet denoising in automotive fmcw radar systems,'' \emph{IEEE Transactions on Intelligent Transportation Systems}, vol.~22, no.~2, pp. 887--897, 2019.

\bibitem{fuchs2020automotive}
J.~Fuchs, A.~Dubey, M.~L{\"u}bke, R.~Weigel, and F.~Lurz, ``Automotive radar interference mitigation using a convolutional autoencoder,'' in \emph{2020 IEEE International Radar Conference (RADAR)}.\hskip 1em plus 0.5em minus 0.4em\relax IEEE, 2020, pp. 315--320.

\bibitem{wang2017automatic}
C.~Wang, J.~Wang, and X.~Zhang, ``Automatic radar waveform recognition based on time-frequency analysis and convolutional neural network,'' in \emph{2017 IEEE International Conference on Acoustics, Speech and Signal Processing (ICASSP)}.\hskip 1em plus 0.5em minus 0.4em\relax IEEE, 2017, pp. 2437--2441.

\bibitem{kong2018automatic}
S.-H. Kong, M.~Kim, L.~M. Hoang, and E.~Kim, ``Automatic lpi radar waveform recognition using cnn,'' \emph{Ieee Access}, vol.~6, pp. 4207--4219, 2018.

\bibitem{zhang2019automatic}
Z.~Zhang, C.~Wang, C.~Gan, S.~Sun, and M.~Wang, ``Automatic modulation classification using convolutional neural network with features fusion of spwvd and bjd,'' \emph{IEEE Transactions on Signal and Information Processing over Networks}, vol.~5, no.~3, pp. 469--478, 2019.

\bibitem{du2022dncnet}
M.~Du, P.~Zhong, X.~Cai, and D.~Bi, ``Dncnet: Deep radar signal denoising and recognition,'' \emph{IEEE Transactions on Aerospace and Electronic Systems}, vol.~58, no.~4, pp. 3549--3562, 2022.

\bibitem{goodfellow2014generative}
I.~Goodfellow, J.~Pouget-Abadie, M.~Mirza, B.~Xu, D.~Warde-Farley, S.~Ozair, A.~Courville, and Y.~Bengio, ``Generative adversarial nets,'' \emph{Advances in neural information processing systems}, vol.~27, 2014.

\bibitem{mvuh2024multichannel}
F.~L. Mvuh, C.~O.~V. Ebode~Ko’a, and B.~Bodo, ``Multichannel high noise level ecg denoising based on adversarial deep learning,'' \emph{Scientific Reports}, vol.~14, no.~1, p. 801, 2024.

\bibitem{kiranyaz2021self}
S.~Kiranyaz, J.~Malik, H.~B. Abdallah, T.~Ince, A.~Iosifidis, and M.~Gabbouj, ``Self-organized operational neural networks with generative neurons,'' \emph{Neural Networks}, vol. 140, pp. 294--308, 2021.

\bibitem{zahid2021robust}
M.~U. Zahid, S.~Kiranyaz, T.~Ince, O.~C. Devecioglu, M.~E. Chowdhury, A.~Khandakar, A.~Tahir, and M.~Gabbouj, ``Robust r-peak detection in low-quality holter ecgs using 1d convolutional neural network,'' \emph{IEEE Transactions on Biomedical Engineering}, vol.~69, no.~1, pp. 119--128, 2021.

\bibitem{gabbouj2022robust}
M.~Gabbouj, S.~Kiranyaz, J.~Malik, M.~U. Zahid, T.~Ince, M.~E. Chowdhury, A.~Khandakar, and A.~Tahir, ``Robust peak detection for holter ecgs by self-organized operational neural networks,'' \emph{IEEE Transactions on Neural Networks and Learning Systems}, vol.~34, no.~11, pp. 9363--9374, 2022.

\bibitem{zahid2022global}
M.~U. Zahid, S.~Kiranyaz, and M.~Gabbouj, ``Global ecg classification by self-operational neural networks with feature injection,'' \emph{IEEE Transactions on Biomedical Engineering}, vol.~70, no.~1, pp. 205--215, 2022.

\bibitem{ince2022blind}
T.~Ince, S.~Kiranyaz, O.~C. Devecioglu, M.~S. Khan, M.~Chowdhury, and M.~Gabbouj, ``Blind restoration of real-world audio by 1d operational gans,'' \emph{arXiv preprint arXiv:2212.14618}, 2022.

\bibitem{kiranyaz2022blind}
S.~Kiranyaz, O.~C. Devecioglu, T.~Ince, J.~Malik, M.~Chowdhury, T.~Hamid, R.~Mazhar, A.~Khandakar, A.~Tahir, T.~Rahman \emph{et~al.}, ``Blind ecg restoration by operational cycle-gans,'' \emph{IEEE Transactions on Biomedical Engineering}, vol.~69, no.~12, pp. 3572--3581, 2022.

\bibitem{zahid2024brsr}
M.~U. Zahid, S.~Kiranyaz, A.~Yildirim, and M.~Gabbouj, ``Brsr-opgan: Blind radar signal restoration using operational generative adversarial network,'' \emph{arXiv preprint arXiv:2407.13949}, 2024.

\bibitem{ahishali2024r2c}
M.~Ahishali, A.~Degerli, S.~Kiranyaz, T.~Hamid, R.~Mazhar, and M.~Gabbouj, ``R2c-gan: Restore-to-classify generative adversarial networks for blind x-ray restoration and covid-19 classification,'' \emph{Pattern Recognition}, vol. 156, p. 110765, 2024.

\bibitem{glorot2010understanding}
X.~Glorot and Y.~Bengio, ``Understanding the difficulty of training deep feedforward neural networks,'' in \emph{Proceedings of the thirteenth international conference on artificial intelligence and statistics}.\hskip 1em plus 0.5em minus 0.4em\relax JMLR Workshop and Conference Proceedings, 2010, pp. 249--256.

\bibitem{loshchilov2016sgdr}
I.~Loshchilov and F.~Hutter, ``Sgdr: Stochastic gradient descent with warm restarts,'' \emph{arXiv preprint arXiv:1608.03983}, 2016.

\bibitem{akiba2019optuna}
T.~Akiba, S.~Sano, T.~Yanase, T.~Ohta, and M.~Koyama, ``{O}ptuna: A next-generation hyperparameter optimization framework,'' in \emph{The 25th ACM SIGKDD International Conference on Knowledge Discovery \& Data Mining}, 2019, pp. 2623--2631.

\bibitem{malik2020fastonn}
J.~Malik, S.~Kiranyaz, and M.~Gabbouj, ``Fastonn--python based open-source gpu implementation for operational neural networks,'' \emph{arXiv preprint arXiv:2006.02267}, 2020.

\end{thebibliography}

\end{document}